\def\paperlanguage{}
\pdfoutput=1 

\documentclass[letterpaper, 10 pt, conference]{ieeeconf}  

\usepackage{bm}
\usepackage{cite}
\usepackage{flushend}
\include{preamble}

\IEEEoverridecommandlockouts                              

\title{\LARGE \textbf
  {
    \switchlanguage%
    {%
      Online Learning Feedback Control Considering Hysteresis\\for Musculoskeletal Structures
    }%
    {%
      筋骨格ヒューマノイドにおけるヒステリシスを考慮した\\オンライン学習型関節フィードバック制御
    }%
  }
}

\author{Kento Kawaharazuka$^{1}$, Kei Okada$^{1}$,  and Masayuki Inaba$^{1}$
  \thanks{$^{1}$ The authors are with the Department of Mechano-Informatics, Graduate School of Information Science and Technology, The University of Tokyo, 7-3-1 Hongo, Bunkyo-ku, Tokyo, 113-8656, Japan.
    {\texttt\small [kawaharazuka, k-okada, inaba]@jsk.t.u-tokyo.ac.jp}
  }
}
\begin{document}

\maketitle
\thispagestyle{empty}
\pagestyle{empty}

\begin{abstract}
  \switchlanguage%
  {%
    While the musculoskeletal humanoid has various biomimetic benefits, its complex modeling is difficult, and many learning control methods have been developed.
    However, for the actual robot, the hysteresis of its joint angle tracking is still an obstacle, and realizing target posture quickly and accurately has been difficult.
    Therefore, we develop a feedback control method considering the hysteresis.
    To solve the problem in feedback controls caused by the closed-link structure of the musculoskeletal body, we update a neural network representing the relationship between the error of joint angles and the change in target muscle lengths online, and realize target joint angles accurately in a few trials.
    We compare the performance of several configurations with various network structures and loss definitions, and verify the effectiveness of this study on an actual musculoskeletal humanoid, Musashi.
  }%
  {%
    筋骨格ヒューマノイドは様々な生物規範型の利点を有すると同時に, そのモデリングが困難であり, 多くの学習型制御手法が開発されてきた.
    しかし, 依然としてヒステリシスが問題となり, 正確な関節角度を実現することは困難だった.
    そこで本研究では, ヒステリシスを考慮した関節フィードバック制御を開発する.
    筋骨格身体の拮抗構造に由来するフィードバックにおける問題を解決するため, 関節角度誤差と指令筋長変化の関係を表すニューラルネットワークをオンラインで学習し, 少数の施行で正確な関節角度を実現する.
    ネットワーク構造や計算方法の違いにによる様々な制御器を構築し, それらの性能を実機において比較実験し, 本研究の有効性を確かめた.
  }%
\end{abstract}

\section{INTRODUCTION}\label{sec:introduction}
\switchlanguage%
{%
  The musculoskeletal humanoid \cite{wittmeier2013toward, jantsch2013anthrob, asano2016kengoro} has various biomimetic benefits such as the redundancy of its muscles \cite{kawaharazuka2022additional, kawaharazuka2022redundancy}, variable stiffness mechanism with nonlinear elastic elements \cite{kawaharazuka2019longtime}, ball joints without extreme points, and the under-actuated spine \cite{asano2016kengoro}.
  On the other hand, applying conventional control methods is unrealistic, because its complex body structure is difficult to modelize.

  Therefore, several learning control methods have been developed so far.
  Mizuuchi, et al. have constructed a neural network which represents the relationship between joint angles and muscle lengths (joint-muscle mapping, JMM) from motion capture data, and realized target joint angles accurately \cite{mizuuchi2006acquisition}.
  Ookubo, et al. have expressed JMM by polynomials, and used it for the state estimation and control \cite{ookubo2015learning}.
  Kawaharazuka, et al. have developed a learning method of JMM by a neural network using vision \cite{kawaharazuka2018online} and its extended method considering the effect of muscle tensions \cite{kawaharazuka2018bodyimage}.
  However, because these methods use a one-to-one relationship between joint angles and muscle lengths, they cannot consider hysteresis of joint angle tracking, which makes realizing target joint angles accurately difficult.

  To solve this problem, feedback control methods are considered.
  Mizuuchi, et al. have developed not only the feedforward but also the feedback control method of joint angles by the change in muscle lengths \cite{mizuuchi2006acquisition}.
  Motegi, et al. have developed a feedback control method of end-effector position by constructing a data table relating the end-effector position with muscle lengths \cite{motegi2012jacobian}.
  However, when executing a feedback control at the fast frequency with the wrong muscle Jacobian, muscles can loosen or break due to high internal muscle tensions caused by its closed link structure.
  To avoid these problems, it is required to complete the feedback trial quickly or estimate its muscle Jacobian accurately.
  Although there are several methods estimating muscle Jacobian for musculoskeletal humanoids \cite{sapio2006shoulder, jantsch2015adaptive, kawaharazuka2018bodyimage}, there will always be some errors.
  Also, although there are methods to realize target posture using reinforcement learning \cite{diamond2014reaching}, they are mostly performed in simulation only and are difficult to handle actual musculoskeletal humanoids with multiple degrees of freedom.

  Therefore, we develop an online learning feedback control method (OLFC) considering hysteresis based on a muscle length-based control for the actual musculoskeletal humanoid.
  To realize target joint angles accurately in a few trials, we update a neural network representing the relationship between the error of joint angles and the change in target muscle lengths online and make use of it for the feedback control.
  This study can consider the time series motion transition of the actual musculoskeletal humanoid, as compared with the previous works of \cite{mizuuchi2006acquisition, kawaharazuka2018online, kawaharazuka2018bodyimage, motegi2012jacobian}.
  Then, we compare the behaviors of OLFC using several experimental settings, network configurations, and loss definitions, in the actual robot experiment.

}%
{%
  筋骨格ヒューマノイド\cite{nakanishi2013design, wittmeier2013toward, jantsch2013anthrob, asano2016kengoro}は筋の冗長性\cite{kobayashi1998tendon}や非線形弾性要素による可変剛性機構\cite{koganezawa1999stiffness}, 特異点のない球関節や劣駆動で多自由度な背骨・指\cite{mizuuchi2004kenta}等の生物規範型の利点を有する.
  同時に, その複雑でモデリング困難な身体構造により, 既存の制御手法を適用することが困難である.

  そこで, これまで多くの学習型制御手法が開発されてきた.
  水内らはモーションキャプチャから得た関節角度と筋長を対応付けたニューラルネットワークを構築し, 指令関節角度をより正確に実現した\cite{mizuuchi2006acquisition}.
  大久保らはIMUから得た関節角度と筋長の関係を多項式近似により表し, 制御や姿勢推定に使用している\cite{ookubo2015learning}.
  河原塚らは視覚から得た関節角度と筋長を対応付けたニューラルネットワークを更新する手法\cite{kawaharazuka2018online}, また, 筋張力の影響も考慮した手法\cite{kawaharazuka2018bodyimage}を開発している.
  しかし, これらの手法は関節角度と筋長の関係を一対一で記述するためヒステリシスを考慮しておらず, 正確な関節角度をフィードフォワードに実現することは難しかった.

  その解決にはフィードバック型の制御が考えられる.
  水内らはフィードフォワード制御に加えて, 筋長変化による関節フィードバック制御を提案している\cite{mizuuchi2006acquisition}.
  また茂木らは, 手先位置と筋長を対応付けるテーブルを構築し, 正確な手先位置を実現する手法を提案している\cite{motegi2012jacobian}.
  ここで問題となるのが, 筋のモーメントアームが一定なモデル化容易な腱駆動ロボット\cite{kobayashi1998tendon, niiyama2010athlete}と違い, 筋骨格ヒューマノイドの関節と筋の関係は正確に算出することが難しいという点である.
  さらに筋骨格構造は閉構造を有しているため, 推定したモーメントアームに誤差がある状態で速い周期でフィードバックを行うと筋が大きく緩む, または筋が強く拮抗し破損するということが起きる.
  そのため, 少数の施行でフィードバックを完了させる, または誤差のない筋のモーメントアームを算出する必要がある.
  他にも, 強化学習\cite{diamond2014reaching}やneural control\cite{richter2016neural, driess2018sqp}により目的の位置に手先を移動させる手法等も存在するが, 実機における実行は難しく, 実機では単関節や2次元平面内の動作に限られる.

  本研究では, 筋長制御によるヒステリシスを考慮した学習型関節フィードバック制御を開発する.
  少数の施行で正確に指令関節角度を実現するため, 関節角度誤差と指令筋長変化の関係を表すニューラルネットワークをオンラインで学習し, フィードバック制御に利用する.
  \secref{sec:basic}では, 筋骨格ヒューマノイドの基本構造, 筋骨格構造におけるヒステリシス特性について説明する.
  また, 水内らが提案した手法\cite{mizuuchi2006acquisition}と河原塚らの手法\cite{kawaharazuka2018bodyimage}を組み合わせて実験を行い, 筋長変化による関節フィードバック制御の特性について考察する.
  \secref{sec:proposed}では, 本研究で提案する筋長変化による関節角度遷移を記述したネットワークのオンライン学習による学習型フィードバック制御について述べる.
  また, 慣性センサや筋張力を考慮したネットワーク, 入出力関係を変化させたネットワークについても述べる.
  \secref{sec:experiment}では, いくつかの状況設定・いくつかのネットワーク構成における比較実験を行う.
  また, オンライン学習の定量的な評価, 現実的な状況におけるタスク実行を行い, 本研究の有効性を確認する.
  最後に, これら手法・実験について考察と結論を述べる.
}%

\begin{figure}[t]
  \centering
  \includegraphics[width=0.65\columnwidth]{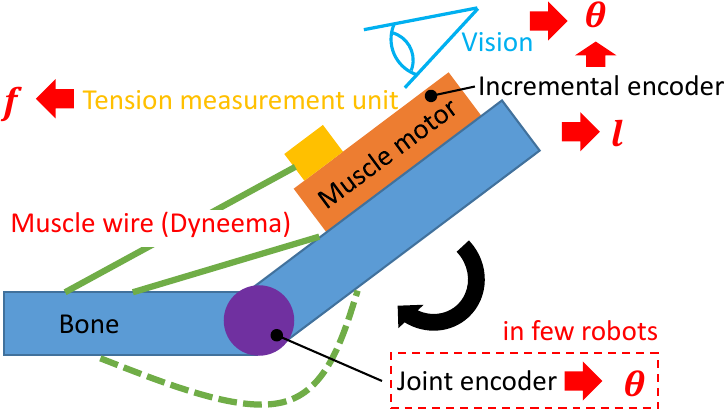}
  \vspace{-1.0ex}
  \caption{Basic structure of the musculoskeletal humanoid.}
  \label{figure:musculoskeletal-structure}
  \vspace{-3.0ex}
\end{figure}

\section{Characteristics of the Musculoskeletal Humanoid and its Basic Feedback Control} \label{sec:basic}
\subsection{Basic Structure of the Musculoskeletal Humanoid} \label{subsec:basic-structure}
\begin{figure}[t]
  \centering
  \includegraphics[width=0.95\columnwidth]{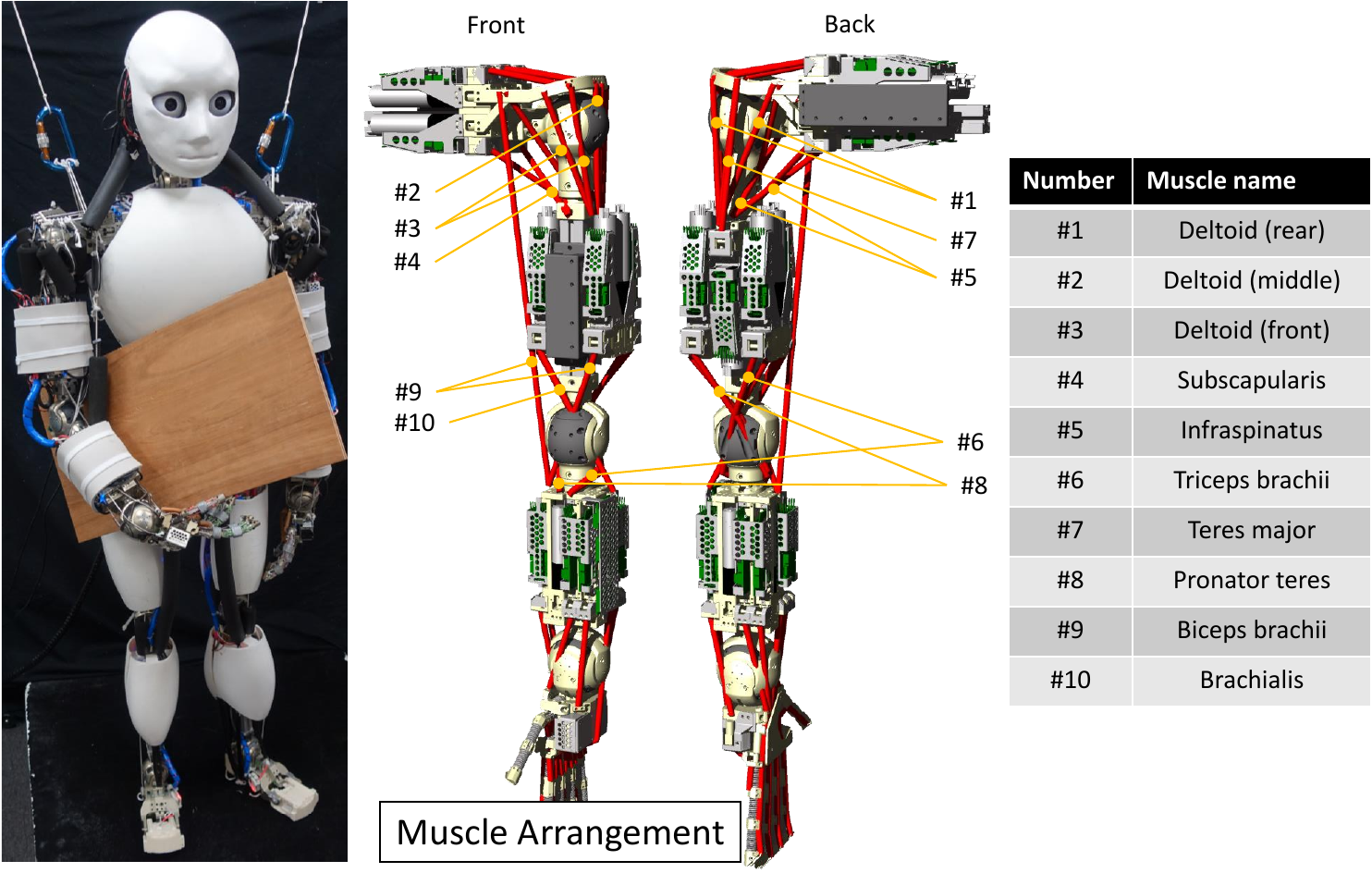}
  \vspace{-1.0ex}
  \caption{A musculoskeletal humanoid Musashi \cite{kawaharazuka2019musashi} used in this study and the muscle arrangement of its left arm.}
  \label{figure:musculoskeletal-humanoid}
  \vspace{-1.0ex}
\end{figure}

\switchlanguage%
{%
  We show the basic body structure of the musculoskeletal humanoid in \figref{figure:musculoskeletal-structure}.
  Muscles are wound by pulleys using electrical motors and antagonistically arranged around joints.
  Muscle length $\bm{l}$ and tension $\bm{f}$ can be measured by encoders and loadcells, respectively.
  The abrasion resistant synthetic fiber Dyneema is used for the muscle wire, and it is slightly elastic.
  Depending on the robot, nonlinear elastic elements are attached to the endpoints of muscles to enable the variable stiffness control.
  Joint angles $\bm{\theta}$ can be directly measured in few robots \cite{kawaharazuka2019musashi, urata2006sensor}.
  However, even if the robot has no joint angle sensors due to complex joints such as the scapula and shoulder ball joint, we can estimate the actual joint angles from changes in muscle lengths and vision information \cite{kawaharazuka2018online}.

  The musculoskeletal humanoid Musashi \cite{kawaharazuka2019musashi} used in this study is shown in the left figure of \figref{figure:musculoskeletal-humanoid} and the muscle arrangement of its left arm is shown in the right figure of \figref{figure:musculoskeletal-humanoid}.
  Musashi has nonlinear elastic elements in muscles and joint angle sensors in joints to investigate and evaluate learning control systems.
  In this study, we mainly use the 3 DOF shoulder and 2 DOF elbow of the left arm of Musashi for experiments.
  We express these joint angles as $\bm{\theta}=(\theta_{S-r}, \theta_{S-p}, \theta_{S-y}, \theta_{E-p}, \theta_{E-y})$ ($S$ means the shoulder, $E$ means the elbow, and $rpy$ means the roll, pitch, and yaw angles, respectively).
  These 5 DOFs include 10 muscles ($\#1$ -- $\#10$ in \figref{figure:musculoskeletal-humanoid}).
  Therefore, in experiments, $\bm{f}$ and $\bm{l}$ are 10 dimensional vectors, and $\bm{\theta}$ is a 5 dimensional vector.
}%
{%
  本研究で扱う筋骨格ヒューマノイドの基本的な構造を\figref{figure:musculoskeletal-humanoid}の左下図に示す.
  関節周りに対して筋が拮抗に冗長配置されており, 筋の筋長と筋張力, 筋温度を測定することができる.
  筋アクチュエータは空気圧等ではなく, 電気モータを使用している.
  筋は摩擦に強い化学繊維であるダイニーマを使用しており, 筋自体が多少伸びる.
  ロボットによっては筋の末端に可変剛性制御を可能とする非線形弾性要素が配置されていることがある.
  また, ロボットによっては関節角度が測定できるように工夫されており, 関節角度が測定できない場合も筋長変化と視覚を用いることで実機関節角度を推定することができる\cite{kawaharazuka2018online}.
  その他ルートリンクに慣性センサ, 手先に接触センサ等が存在する場合もある.

  本研究で使用する筋骨格ヒューマノイドMusashi\cite{kawaharazuka2019musashi}は, 筋に非線形弾性要素を, 関節には学習型制御模索のために関節角度センサを有している.
  本研究では主に, 肩の3自由度, 肘の2自由度を用いて実験を行うこととし, 関節角度は$\bm{\theta}=(\theta_{S-r}, \theta_{S-p}, \theta_{S-y}, \theta_{E-p}, \theta_{E-y})$のように表す($S$はshoulder, $E$はelbow, $rpy$はそれぞれroll, pitch, yawを表す).
}%

\begin{figure}[t]
  \centering
  \includegraphics[width=0.65\columnwidth]{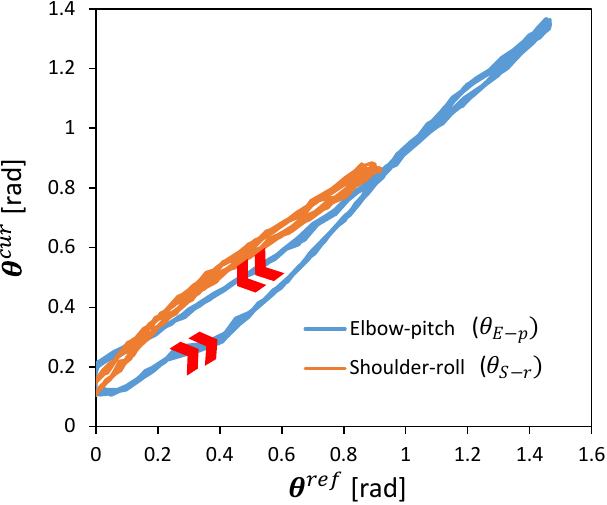}
  \vspace{-1.0ex}
  \caption{Characteristics of hysteresis in the musculoskeletal structure.}
  \label{figure:muscle-hysteresis}
  \vspace{-3.0ex}
\end{figure}

\subsection{Characteristics of Hysteresis in the Musculoskeletal Structure} \label{subsec:hysteresis}
\switchlanguage%
{%
  We conducted two experiments to investigate the characteristics of hysteresis in the musculoskeletal structure.
  First, we examined the tracking of the target joint angle $\theta^{ref} = \theta_{E-p}$ by changing it from 0 to -90 [deg] and from -90 to 0 [deg] over 10 sec and repeating this 10 times (we initialized the robot posture by $(\theta_{S-r}, \theta_{S-p}, \theta_{S-y}, \theta_{E-p}, \theta_{E-y}) = (0, 0, 0, 0, 0)$ [deg]).
  Second, we likewise examined the tracking of $\theta^{ref} = \theta_{S-r}$ by changing it from 0 to 60 [deg] and from 60 to 0 [deg] over 10 sec and repeating this 10 times.
  In these experiments, we converted the target joint angle $\bm{\theta}^{ref}$ to the target muscle length $\bm{l}^{ref}$ using \cite{kawaharazuka2018bodyimage}, and sent it to the actual robot.

  We show the transition of $\bm{\theta}^{ref}$ and the current joint angle $\bm{\theta}^{cur}$ in \figref{figure:muscle-hysteresis}.
  The graph of $\theta_{E-p}$ is turned over for better visual understandability.
  From these results, hysteresis exists in the joint angle movements.
  Its characteristics depend on joints such as $S-r$ and $E-p$, and the hysteresis has reproducibility because there is almost no difference among the 10 repeated movements.
}%
{%
  筋骨格ヒューマノイドにおけるヒステリシス特性について2つの実験を行った.
  まず, 初期姿勢$(\theta_{S-r}, \theta_{S-p}, \theta_{S-y}, \theta_{E-p}, \theta_{E-y}) = (0, 0, 0, 0, 0)$ [deg]の状態から, $\theta_{E-p}$の指令値$\theta^{ref}$を10 secで-90 degまで動かし, 元に戻すことを10回行ったときの実機関節角度$\theta^{cur}$の追従を調べる.
  次に, 同様に初期姿勢から$\theta_{S-r}$の指令値$\theta^{ref}$を10 secで60 degまで動かし, 元に戻すことを10回行ったときの実機関節角度$\theta^{cur}$の追従を調べる.
  このとき, 指令関節角度$\bm{\theta}^{ref}$は\cite{kawaharazuka2018bodyimage}により実機学習されたネットワーク$\bm{h}(\bm{\theta}^{ref}, \bm{f}^{ref})$を用いて指令筋長$\bm{l}^{ref}$に変換され実機に送られる(ここで, $\bm{f}^{ref}$は指令筋張力を表す).
  このネットワーク$\bm{h}$は, ある$\bm{\theta}^{ref}$, $\bm{f}^{ref}$を実現する$\bm{l}^{ref}$を出力するネットワークである.
  ただし, 本節では$\bm{f}^{ref} = \bm{f}_{const}$ ($\bm{f}_{const}$はある一定の筋張力, 本研究では30 N)とする.

  実験結果を\figref{figure:muscle-hysteresis}に示す.
  ここでは, わかりやすいよう$\theta_{E-p}$は反転している.
  グラフから, 関節角度制御に対してヒステリシスがあることがわかる.
  その特性は$S-r$や$E-p$等の関節に依存して異なり, また, 10回の動作でほとんど差がないため, 再現性は高いこともわかる.
  これは, 筋のダイニーマのヒステリシス, 筋の経由点における摩擦, 関節の摩擦等, 様々な要因が考えられる.
}%

\subsection{Basic Feedback Control} \label{subsec:basic}
\switchlanguage%
{%
  As the simplest way to solve the hysteresis problem, we show the basic feedback control method (BFC) \cite{mizuuchi2006acquisition}.
  We execute BFC using not the original network $\bm{h}_{orig}(\bm{\theta})$ in \cite{mizuuchi2006acquisition} but the latest network $\bm{h}(\bm{\theta}, \bm{f})$ in \cite{kawaharazuka2018online}, which can be trained using the actual robot sensor information.
  In BFC, the target change in muscle length $\Delta\bm{l}$, when given the current joint angle $\bm{\theta}^{cur}$ and the target joint angle $\bm{\theta}^{ref}$, is calculated as below,
  \begin{align}
    \Delta\bm{l}(\bm{\theta}^{ref}&, \bm{\theta}^{cur}) = \bm{h}(\bm{\theta}^{ref},\bm{f}^{cur})-\bm{h}(\bm{\theta}^{cur}, \bm{f}^{cur}) \label{eq:basic}
  \end{align}
  where $\bm{f}^{cur}$ is the current muscle tension.
  In \equref{eq:basic}, we assume that $\bm{\theta}^{cur}$ is close to $\bm{\theta}^{ref}$ and so $\bm{f}^{cur}$ is not largely changed by sending $\Delta\bm{l}$.
  $\bm{\theta}^{ref}$ is realized to some extent by $\bm{l}^{ref} = \bm{h}(\bm{\theta}^{ref}, \bm{f}_{const})$ first ($\bm{f}_{const}$ is a certain constant value; 30 N in this study), and $\bm{\theta}^{ref}$ is accurately realized by repeating \equref{eq:basic} as $\bm{l}^{ref} \gets \bm{l}^{ref} + \Delta\bm{l}$.
}%
{%
  ヒステリシスを解決するための制御として, \cite{mizuuchi2006acquisition}で提案されたフィードバック制御(BFC)を以下に示す.
  ただし, \cite{mizuuchi2006acquisition}において用いられたJMM $\bm{h}_{orig}(\bm{\theta})$ではなく, より最新の研究に基づき, \secref{subsec:hysteresis}でも利用した\cite{kawaharazuka2018bodyimage}において実機データを用いて訓練されるJMM, $\bm{h}(\bm{\theta}, \bm{f})$を用いてこれを行う.
  現在関節角度$\bm{\theta}^{cur}$と指令関節角度$\bm{\theta}^{ref}$を与えたときのフィードバックによる筋長指令の変化項$\Delta\bm{l}$は以下のようになる.
  \begin{align}
    \Delta\bm{l}(\bm{\theta}^{ref}&, \bm{\theta}^{cur}, \bm{f}^{cur})\nonumber\\
    &= \bm{h}(\bm{\theta}^{ref},\bm{f}^{cur})-\bm{h}(\bm{\theta}^{cur}, \bm{f}^{cur}) \label{eq:basic}
  \end{align}
  ここで, $\bm{f}^{cur}$は現在の筋張力を表す.
  \equref{eq:basic}では$\bm{\theta}^{cur}$と$\bm{\theta}^{ref}$はある程度近く, そのため$\bm{f}$は大きくは変化しないという仮定を置いている.
  実際にある$\bm{\theta}^{ref}$を実現する際は, $\bm{l}^{ref} = \bm{h}(\bm{\theta}^{ref}, \bm{f}_{const})$により指令関節角度に近い関節角度を実現し, その後\equref{eq:basic}を繰り返すことにより正確に$\bm{\theta}^{ref}$を実現する, という手順を取る.
}%

\begin{figure}[t]
  \centering
  \includegraphics[width=0.8\columnwidth]{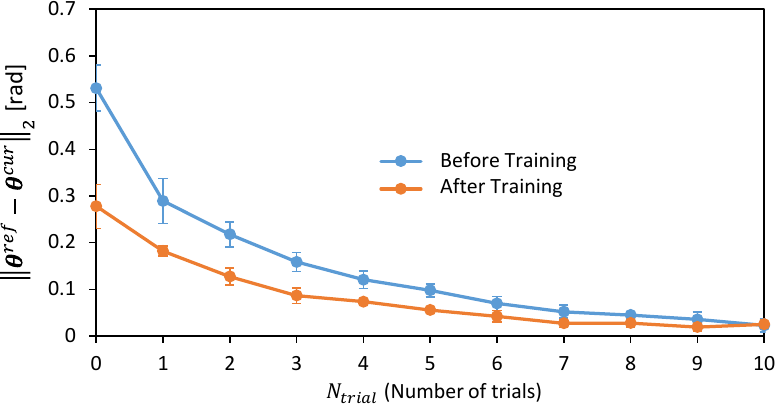}
  \vspace{-1.0ex}
  \caption{Evaluation experiment of the basic feedback control using a network of \cite{kawaharazuka2018bodyimage}. This graph includes the results before and after the training of the network using the actual robot sensor information.}
  \label{figure:basic-evaluation}
  \vspace{-3.0ex}
\end{figure}

\subsection{The Evaluation of Basic Feedback Control} \label{subsec:basic-eval}
\switchlanguage%
{%
  We evaluated BFC in an actual robot experiment by repeating the $N^{max}_{trial}$ times execution of \equref{eq:basic} $N^{max}_{rand}$ times.
  We conducted an experiment to realize the target joint angle $\bm{\theta}^{ref}$ of $(\theta_{S-r}, \theta_{S-p}, \theta_{S-y}, \theta_{E-p}, \theta_{E-y}) = (30, -30, 30, -60, 30)$ [deg] from 5 randomly chosen joint angles $\bm{\theta}_{rand}$.
  First, the target joint angle is realized by $\bm{l}^{ref} = \bm{h}(\bm{\theta}^{ref}, \bm{f}_{const})$ to some extent.
  Then, the trial is set as $N_{trial}=0$, and $||\bm{\theta}^{ref}-\bm{\theta}^{cur}||_{2}$ ($||\cdot||_{2}$ expresses L2 norm) is measured every time \equref{eq:basic} is performed.
  \equref{eq:basic} is quasi-statically executed over 3 sec.
  By repeating \equref{eq:basic} $N^{max}_{trial}$ times ($0 \leq N_{trial} < N^{max}_{trial}$) from $\bm{\theta}^{cur}$ to $\bm{\theta}^{ref}$ while changing the initial posture $\bm{\theta}_{rand}$ $N^{max}_{rand}$ times ($0 \leq N_{rand} < N^{max}_{rand}$), we check the average and variance of $||\bm{\theta}^{ref}-\bm{\theta}^{cur}||_{2}$ of $N^{max}_{rand}$ times at the $N_{trial}$-th trial.
  We set $N^{max}_{trial}=10$ and $N^{max}_{rand}=5$ in this experiment.

  The graph in \figref{figure:basic-evaluation} shows the result of conducting \equref{eq:basic} using $\bm{h}$, before and after its training using the actual robot sensor information in \cite{kawaharazuka2018bodyimage}.
  After the training of $\bm{h}$, $||\bm{\theta}^{ref}-\bm{\theta}^{cur}||_{2}$ at $N_{trial}=0$ drops by half compared to the result before the training of $\bm{h}$.
  Although the final results of $||\bm{\theta}^{ref}-\bm{\theta}^{cur}||_{2}$ do not largely vary, the convergence is faster after the training of $\bm{h}$ than before it.
  However, both require many trials to converge, and the current learning method training one-to-one relationship between joint angles and muscle lengths is not enough.

  The method of BFC does not consider hysteresis.
  Thus, $\Delta\bm{l}(\bm{\theta}_{1}, \bm{\theta}_{2})=\Delta\bm{l}(\bm{\theta}_{2}, \bm{\theta}_{1})$, when setting the random joint angles of $\bm{\theta}_{1}$ and $\bm{\theta}_{2}$.
  However, because hysteresis exists as shown in \secref{subsec:hysteresis}, the target joint angle cannot be easily realized, the number of trials of \equref{eq:basic} increases, and muscles can loosen or high muscle tension can emerge as stated in \secref{sec:introduction}.
  To realize the target joint angle in a few trials of feedback control, we propose online learning feedback control next.
}%
{%
  前節の手法が有効かを実験により示す.
  ランダムな5つの関節角度$\bm{\theta}_{rand}$から$(\theta_{S-r}, \theta_{S-p}, \theta_{S-y}, \theta_{E-p}, \theta_{E-y}) = (30, -30, 30, -60, 30)$ [deg]の指令関節角度$\bm{\theta}^{ref}$を実現する実験を行う.
  初めに$\bm{l}^{ref} = \bm{h}(\bm{\theta}^{ref}, \bm{f}_{const})$によりある程度関節角度を実現し, それを施行$N_{trial}=0$回目として\equref{eq:basic}を行う度に$||\bm{\theta}^{ref}-\bm{\theta}^{cur}||_{2}$を計測する($||\cdot||_{2}$はL2ノルムを表す).
  準静的に, 一回の\equref{eq:basic}は3 secかけて動作させる.
  $\bm{\theta}^{cur}$から$\bm{\theta}^{ref}$まで$N^{max}_{trial}=10$回($0 \leq N_{trial} < N^{max}_{trial}$)\equref{eq:basic}を行うことを, 初期関節角度$\bm{\theta}_{rand}$を変えながら$N^{max}_{rand}=5$回($0 \leq N_{rand} < N^{max}_{rand}$)繰り返し, $N_{trial}$回目における$N^{max}_{rand}$回分の$||\bm{\theta}^{ref}-\bm{\theta}^{cur}||_{2}$の平均と分散の遷移を確認する.

  結果は\figref{figure:basic-evaluation}のようになっている.
  \cite{kawaharazuka2018bodyimage}により学習される前と学習された後の$\bm{h}$を用いて\equref{eq:basic}を行った際の結果を示している.
  \cite{kawaharazuka2018bodyimage}により学習された後は, 学習される前に比べて$N_{trial}=0$で$||\bm{\theta}^{ref}-\bm{\theta}^{cur}||_{2}$が半分程度になっている.
  最終的な収束結果はあまり変わらないが, 全体として\cite{kawaharazuka2018bodyimage}による学習後の方が収束は速い.
  しかし, 両者とも収束までは多くの施行を必要とする.

  本章で述べた方法はヒステリシスが考慮されていない.
  つまり, あるランダムな$\bm{\theta}_{1}$と$\bm{\theta}_{2}$を指定したとき, $\Delta\bm{l}(\bm{\theta}_{1}, \bm{\theta}_{2})=\Delta\bm{l}(\bm{\theta}_{2}, \bm{\theta}_{1})$となる.
  しかし, 実際には\secref{subsec:hysteresis}にあるように, ヒステリシスが存在するため簡単に$\bm{\theta}^{ref}$を実現することはできず, フィードバックの施行回数$N_{trial}$が重なり, \secref{sec:introduction}で説明したように筋の緩みや内力の高まりが起きる可能性がある.
  より少ない施行でのフィードバックを目指すためにはこれらモデル化の難しいヒステリシスの影響を学習的に考慮していく必要がある.
}%

\section{Online Learning Feedback Control} \label{sec:proposed}
\switchlanguage%
{%
  In this section, we will propose a method to acquire the target change in muscle length $\Delta\bm{l}$ to realize $\bm{\theta}^{ref}$ from $\bm{\theta}^{cur}$ by online learning.
  We will explain two different network structures for the purpose, their initial training and online learning, and feedback controls using the trained networks.
}%
{%
  本章では, \secref{subsec:basic}, \secref{subsec:basic-eval}で述べたような$\bm{\theta}^{cur}$を$\bm{\theta}^{ref}$まで動かすための$\Delta\bm{l}$を, 学習的に獲得していく手法を提案する.
}%

\begin{figure}[t]
  \centering
  \includegraphics[width=0.8\columnwidth]{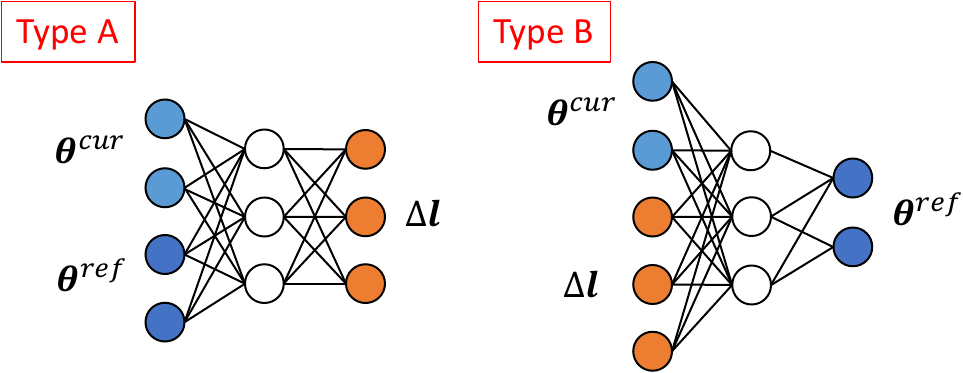}
  \vspace{-1.0ex}
  \caption{Network structures for online learning feedback control.}
  \label{figure:learning-networks}
  \vspace{-1.0ex}
\end{figure}

\begin{figure}[t]
  \centering
  \includegraphics[width=0.95\columnwidth]{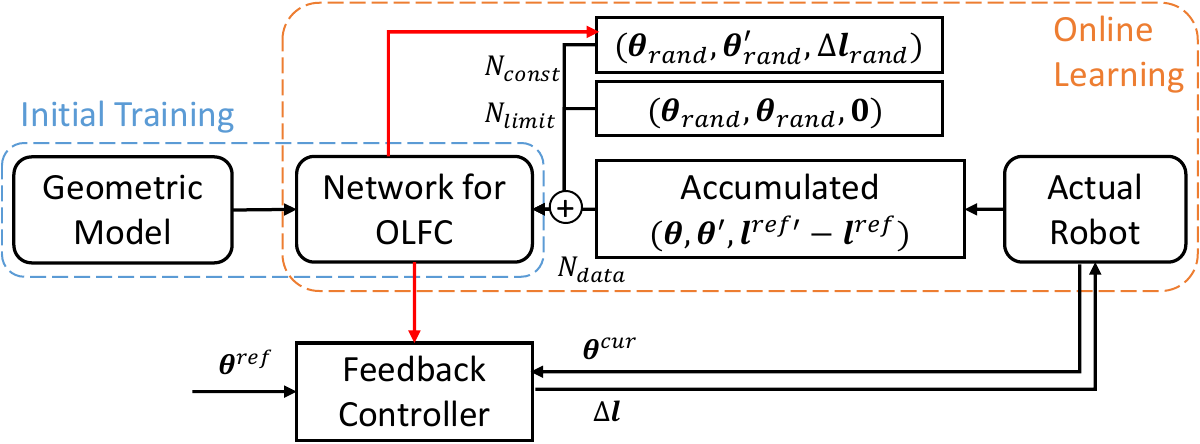}
  \vspace{-1.0ex}
  \caption{The overall system of this study.}
  \label{figure:system}
  \vspace{-3.0ex}
\end{figure}

\subsection{Network Structures} \label{subsec:network-structure}
\switchlanguage%
{%
  We can consider two neural network structures for online learning feedback control (OLFC) shown in \figref{figure:learning-networks}.
  These structures are chosen by considering all combinations of $\bm{\theta}^{cur}$, $\bm{\theta}^{ref}$, and $\Delta\bm{l}$ as network input and output, and only the chosen two networks are valid.
  First, the simplest network structure to realize our objective ($\bm{h}_{A}$, Type A) is represented as below.
  \begin{align}
    \Delta\bm{l} = \bm{h}_{A}(\bm{\theta}^{cur}, \bm{\theta}^{ref})
  \end{align}
  By using Type A, $\Delta\bm{l}$ can be directly calculated when given $\bm{\theta}^{cur}$ and $\bm{\theta}^{ref}$.
  Second, when examining the relationship among $\bm{\theta}^{cur}$, $\bm{\theta}^{ref}$, and $\Delta\bm{l}$, we can consider a network structure representing the transition of joint angles with the change in muscle length ($\bm{h}_{B}$, Type B), as below.
  \begin{align}
    \bm{\theta}^{ref} = \bm{h}_{B}(\bm{\theta}^{cur}, \Delta\bm{l})
  \end{align}
  By using a backpropagation \cite{rumelhart1986backprop} and gradient descent method for the inputs, $\Delta\bm{l}$ can also be calculated from $\bm{\theta}^{cur}$ and $\bm{\theta}^{ref}$.
  We show the overall system of this study in \figref{figure:system}.
}%
{%
  ネットワーク構造を2つ\figref{figure:learning-networks}に示す.
  本研究の目的を実現する最も単純なネットワーク構造は, $\bm{\theta}^{cur}$と$\bm{\theta}^{ref}$を入力として, $\Delta\bm{l}$を出力するネットワークであろう (Type A).
  また, $\bm{\theta}^{cur}$と$\Delta\bm{l}$を入力として$\bm{\theta}^{ref}$を出力するような関節角度の遷移を表す状態方程式型のネットワーク構造も考えられる (Type B).

}%

\subsection{Initial Training of the Networks} \label{subsec:initial-training}
\switchlanguage%
{%
  We collect the data of $(\bm{\theta}^{cur}, \bm{\theta}^{ref}, \Delta\bm{l})$ using a geometric model and train the networks.
  The geometric model has information about joint position, link length, and link weight.
  Also, each muscle route is expressed by linearly connecting its start point, relay points, and end point.
  By setting a certain joint angle, muscle length can be calculated from the distance of the start, relay, and end points.
  This geometric model is designed by humans, and cannot consider the muscle wrapping around the skeleton, the elongation of the muscle wire, hysteresis, etc.

  First, we calculate the muscle length $\bm{l}$ at a random joint angle $\bm{\theta}$ from the geometric model.
  Second, we calculate the muscle length $\bm{l}'$ at the joint angle of $\bm{\theta}' = \bm{\theta}+\Delta\bm{\theta}$.
  In this procedure, $\Delta\bm{\theta}$ is a random value following a normal distribution with an average of $0$ and a standard deviation of $S_{\Delta\bm{\theta}}$ ($S_{\Delta\bm{\theta}}=0.5$ [rad] in this study).
  Then, we can construct a dataset of $(\bm{\theta}^{cur}=\bm{\theta}, \bm{\theta}^{ref}=\bm{\theta}', \Delta\bm{l}=\bm{l}'-\bm{l})$ by repeating the above procedure, and train these networks using it.
  In this study, each network has 5 fully-connected layers including the input and output layers, and the numbers of units in the hidden layers are (40, 20, 20).
  Also, the loss function is a mean squared error, the activation function is Sigmoid, and the optimization method is Adam \cite{kingma2015adam}.
}%
{%
  これらネットワークの幾何モデルを用いた初期学習について述べる.
  ここで言う幾何モデルとは, 筋の起始点・中継点・終止点を直線で結び筋経路を表現したものである.

  初めに, 幾何モデルからランダムな関節角度$\bm{\theta}$と対応する筋長$\bm{l}$を計算する.
  その後, $\bm{\theta}' = \bm{\theta}+\Delta\bm{\theta}$と対応する筋長$\bm{l}'$を計算する.
  ここで, $\Delta\bm{\theta}$は平均0, 標準偏差$S_{\Delta\bm{\theta}}$(本研究では, $S_{\Delta\bm{\theta}}=0.5$ [rad])とする.
  よって, $(\bm{\theta}^{cur}=\bm{\theta}, \bm{\theta}^{ref}=\bm{\theta}', \Delta\bm{l}=\bm{l}'-\bm{l})$のデータセットが構成でき, これを用いてネットワークを学習させる.
  本研究では全ネットワークを入力・出力層を含め5層とし, 中間層のユニット数は(40, 20, 20)となっている.
  また, 活性化関数はSigmoid, 最適化はAdam \cite{kingma2015adam}によって行う.

}%

\subsection{Online Learning of the Network} \label{subsec:online-learning}
\switchlanguage%
{%
  First, a joint angle movement from the previous joint angle $\bm{\theta}$ to the current joint angle $\bm{\theta}'$ is detected using the change in target muscle length from $\bm{l}^{ref}$ to $\bm{l}^{ref'}$.
  Then, the data of $(\bm{\theta}^{cur}=\bm{\theta}, \bm{\theta}^{ref}=\bm{\theta}', \Delta\bm{l}=\bm{l}^{ref'}-\bm{l}^{ref})$ can be obtained.
  Because of the assumption that $\bm{\theta}$ is close to $\bm{\theta}'$ as stated in \secref{subsec:basic}, the data is used only when $||\bm{l}^{ref}-\bm{l}^{ref'}||_{2}<C_{length}$ ($C_{length}$ is a threshold constant).
  We repeat this step, accumulate the data, and start the online learning when the number of the accumulated data exceeds a threshold value $N_{thre}$.
  We choose $N_{data}$ ($N_{data}$ is a constant, $N_{data} \leq N_{thre}$) number of data from these accumulated data and add them into a batch for online learning.
  In order to add the limitation of ``$\Delta\bm{l}=\bm{0}$ when $\bm{\theta}^{cur}=\bm{\theta}^{ref'}$, we add $N_{limit}$ number of data $(\bm{\theta}_{rand}, \bm{\theta}_{rand}, \bm{0})$ into the batch by randomly choosing $\bm{\theta}_{rand}$ ($N_{limit}$ is a constant).
  Also, the network should not be changed except for around the obtained data.
  Thus, regarding Type A, we add $\bm{N}_{const}$ number of data into the batch ($\bm{N}_{const}$ is a constant), by calculating $\Delta\bm{l}_{rand}$ from the randomly chosen $\bm{\theta}_{rand}$ and $\bm{\theta}'_{rand}$ using the current network.
  Regarding Type B, we likewise calculate $\bm{\theta}'_{rand}$ from the randomly chosen $\bm{\theta}_{rand}$ and $\Delta\bm{l}_{rand}$.
  Finally, we set the number of epochs as $N_{epoch}$ and the batch size as $N_{data}+N_{limit}+N_{const}$, which is the sum of the data stated above, and train the network.
  In this study, we set $C_{length}=100$ [mm], $N_{thre}=10$, $N_{data}=10$, $N_{limit}=5$, $N_{const}=5$, and $N_{epoch}=3$.
}%
{%
  これらネットワークのオンライン学習方法について述べる.
  まず, 関節角度が静止状態の$\bm{\theta}$から静止状態の$\bm{\theta}$'まで動いたことを, $\bm{l}^{ref}$の変化から検知する.
  ただし, \secref{subsec:basic}における仮定と同様, $\bm{\theta}$と$\bm{\theta}'$は近いという仮定を持つため, $||\bm{l}^{ref}-\bm{l}^{ref'}||_{2}<C_{length}$のときのみをデータとして用いる.
  ここから, $(\bm{\theta}^{cur}=\bm{\theta}, \bm{\theta}^{ref}=\bm{\theta}', \Delta\bm{l})$を取得することができる.
  次にこれを蓄積していき, データ数が$N_{thre}$を超えたところでネットワークのオンライン学習を始める.
  蓄積したデータの中から$N_{data}$個のデータを選ぶ.
  また, $\bm{\theta}=\bm{\theta}'$のとき$\Delta\bm{l}=\bm{0}$という制約のデータセットを, $\bm{\theta}=\bm{\theta}'$をランダムに選び$N_{limit}$個作成する.
  また, 学習されるデータ付近以外はネットワークは変更されるべきではない.
  そこでType Aならば, ランダムに$\bm{\theta}, \bm{\theta}'$を現在のネットワークに入れて$\Delta\bm{l}$を算出したデータを$\bm{N}_{const}$個作成する.
  Type Bも同様であり, ランダムに$\bm{\theta}, \Delta\bm{l}$をネットワークに入れてデータを作成する.
  最後に, 集まった$N_{data}+N_{limit}+N_{const}$個のデータをバッチとして, $N_{epoch}$エポックネットワークを学習させる.
  本研究では, $C_{length}=100$ [mm], $N_{thre}=10$, $N_{data}=10$, $N_{limit}=5$, $N_{const}=5$, $N_{epoch}=3$とする.
}%

\subsection{Feedback Control Using the Learned Network} \label{subsec:learning-feedback}
\switchlanguage%
{%
  The network of Type A is the simplest, and we can obtain $\Delta\bm{l}$ through the network by merely measuring $\bm{\theta}^{cur}$ and deciding $\bm{\theta}^{ref}$.

  On the other hand, several procedures to obtain $\Delta\bm{l}$ are required for the network of Type B.
  First, we calculate the original target change in muscle length $\Delta\bm{l}_{geo}$ using the geometric model as shown below,
  \begin{align}
    \Delta\bm{l}_{geo} = \bm{h}_{geo}(\bm{\theta}^{ref}) - \bm{h}_{geo}(\bm{\theta}^{cur})
  \end{align}
  where $\bm{h}_{geo}$ is the mapping from joint angle to muscle length in the geometric model.
  $\Delta\bm{l}_{geo}$ is almost the same value as the output of \equref{eq:basic} before the training of $\bm{h}$ in \cite{kawaharazuka2018bodyimage}.
  Second, we obtain the predicted joint angle $\bm{\theta}^{pred}$ by feeding $\bm{\theta}^{cur}$ and $\Delta\bm{l}_{geo}$ into the network of Type B.
  Then, we define the loss function $L$ as shown below, and update $\Delta\bm{l}_{geo}$ through backpropagation \cite{rumelhart1986backprop} and descent gradient method,
  \begin{align}
    L_{0} &= ||\bm{\theta}^{pred}-\bm{\theta}^{ref}||_{2} \label{eq:loss0}\\
    \Delta\bm{l}_{geo} &\gets \Delta\bm{l}_{geo}-\gamma\partial{L}/\partial\Delta\bm{l}_{geo}
  \end{align}
  where $\gamma$ is a learning rate.
  Although we can set $\gamma$ as a constant value, we change $\gamma$ variously for better convergence.
  We decide the maximum value of $\gamma$ as $\gamma_{max}$, equally divide $[0, \gamma_{max}]$ into $N_{batch}$ parts, update $\Delta\bm{l}_{geo}$ by each learning rate, and choose the one with the minimum loss $L$ when feeding it into the network again.
  By repeating these procedures and updating $\Delta\bm{l}_{geo}$, $N_{update}$ times ($N_{update}$ is a constant), we can finally obtain the refined $\Delta\bm{l}_{geo}$ as the target $\Delta\bm{l}$.

  This method has some extensions.
  For example, by defining $L$ as shown below and executing the same procedure as stated above, we can calculate $\Delta\bm{l}$ to realize $\bm{\theta}^{ref}$ with the minimum change in muscle length,
  \begin{align}
    L_{1} = ||\bm{\theta}^{pred}-\bm{\theta}^{ref}||_{2} + \alpha||\Delta\bm{l}||_{2} \label{eq:loss1}
  \end{align}
  where $\alpha$ is a weight coefficient.
  Similarly, by defining $L$ as shown below, we can inhibit only the contracting muscles and avoid high internal muscle tension.
  \begin{align}
    \Delta\bm{l}_{processed} =
    \begin{cases}
      \Delta{l}, & if\;\;\Delta{l} < 0\\
      0, & else
    \end{cases}\\
      L_{2} = ||\bm{\theta}^{pred}-\bm{\theta}^{ref}||_{2} + \alpha||\Delta\bm{l}_{processed}||_{2} \label{eq:loss2}
  \end{align}

  In this study, we set $\gamma_{max}=10$ [mm], $N_{batch}=10$, $N_{update}=10$, and $\alpha=0.0003$, and we use \equref{eq:loss0} as the loss function unless otherwise noted.
}%
{%
  学習されたネットワークを用いて関節フィードバック制御を行う方法を述べる.

  Type Aのネットワークは最も単純であり, $\bm{\theta}^{cur}$と$\bm{\theta}^{ref}$を決めて, ネットワークを通すことで$\Delta\bm{l}$を得るのみである.
  それに対して, Type Bのネットワークではいくつかの手順を踏む.
  まず, 幾何モデルを用いて$\bm{\theta}^{cur}$と$\bm{\theta}^{ref}$から\equref{eq:basic}のように$\Delta\bm{l}_{geo}$を求める.
  これは, 学習されていないネットワークにおける出力と同じである.
  次に, $\Delta\bm{l}_{geo}$を初期値として, $\bm{\theta}^{cur}$と伴にネットワークを通し, $\bm{\theta}^{pred}$を得る.
  ここで, 損失$L$を,
  \begin{align}
    L = ||\bm{\theta}^{pred}-\bm{\theta}^{ref}||_{2} \label{eq:loss0}
  \end{align}
  として, 誤差逆伝播\cite{rumelhart1986backprop}により$\Delta\bm{l}_{geo}$を以下のように更新していく.
  \begin{align}
    \bm{g} &= dL/d\Delta\bm{l}_{geo}\\
    \Delta\bm{l}_{geo} &\gets \Delta\bm{l}_{geo}-\gamma\partial{L}/\partial\Delta\bm{l}_{geo}
  \end{align}
  ここで, $\gamma$は学習率を表す.
  $\gamma$を固定で決めても良いが, 本研究では$\gamma$の最大値$\gamma_{max}$を決め, $[0, \gamma_{max}]$を$N_{batch}$等分し, それぞれの学習率で更新した$\Delta\bm{l}_{geo}$をもう一度ネットワークに通して最も$L$が小さかったものを採用する.
  この手順を$N_{update}$回繰り返して$\Delta\bm{l}_{geo}$を更新していき, 最終的に求まった解を$\Delta\bm{l}$とする.

  この手法にはいくつかの発展があり, 例えば$L$を
  \begin{align}
    L = ||\bm{\theta}^{pred}-\bm{\theta}^{ref}||_{2} + \alpha||\Delta\bm{l}||_{2} \label{eq:loss1}
  \end{align}
  とすることで, 最小限の筋長変化で$\bm{\theta}^{ref}$を実現する$\Delta\bm{l}$を計算することができる.
  ここで$\alpha$は重みの係数である.
  同様に, 
  \begin{align}
    \Delta\bm{l}_{processed} =
    \begin{cases}
      \Delta{l}, & if\;\;\Delta{l} < 0\\
      0, & else
    \end{cases}\\
      L = ||\bm{\theta}^{pred}-\bm{\theta}^{ref}||_{2} + \alpha||\Delta\bm{l}_{processed}||_{2} \label{eq:loss2}
  \end{align}
  とすることで, 収縮する筋のみ可能な限り抑制することができ, 拮抗による高い内力を避けることができる.

  本研究では$\gamma_{max}=10$ [mm], $N_{batch}=10$, $N_{update}=10$, $\alpha=0.0003$とし, 断りがない限り$L$としては\equref{eq:loss0}を用いる.
}%

\section{Experiments} \label{sec:experiment}
\switchlanguage%
{%
  First, we will conduct OLFC using the network of Type A and B, and compare the convergence of joint angles between $\bm{\theta}^{cur}$ and $\bm{\theta}^{ref}$ among these networks.
  Second, we will compare the difference in the behavior of OLFC among the different loss definitions and among different situations such as changing the posture or grasping a heavy object.
  Also, we will conduct a quantitative evaluation of OLFC.
  Finally, we will conduct a practical task of Japanese stamp pushing using the proposed method and visual information.

  In this section, we will mainly conduct experiments to realize the target joint angle $\bm{\theta}^{ref}$ from the randomly chosen joint angle $\bm{\theta}_{rand}$, as in \secref{subsec:basic-eval}.
  First, we realize the target joint angle by $\bm{l}^{ref} = \bm{h}(\bm{\theta}^{ref}, \bm{f}_{const})$ to some extent ($\bm{h}$ is already trained).
  Then, setting the trial as $N_{trial}=0$, we measure $||\bm{\theta}^{ref}-\bm{\theta}^{cur}||_{2}$, conducting OLFC stated in \secref{subsec:learning-feedback} every time.
  By repeating OLFC $N^{max}_{trial}$ times ($0 \leq N_{trial} < N^{max}_{trial}$) from $\bm{\theta}^{cur}$ to $\bm{\theta}^{ref}$ while changing the initial posture $\bm{\theta}_{rand}$ $N^{max}_{rand}$ times ($0 \leq N_{rand} < N^{max}_{rand}$), we check the average and variance of $||\bm{\theta}^{ref}-\bm{\theta}^{cur}||_{2}$ of $N^{max}_{rand}$ times at the $N_{trial}$-th trial.
  We express this graph as $\textrm{Graph}_{ave}(\bm{\theta}^{ref}, N^{max}_{trial}, N^{max}_{rand})$.
  Also, $\textrm{Graph}_{raw}(\bm{\theta}^{ref}, N^{max}_{trial}, N^{max}_{rand})$ expresses $N^{max}_{rand}$ times the transition of $||\bm{\theta}^{ref}-\bm{\theta}^{cur}||_{2}$ without averaging them.
  Here, all constants are defined as with the preliminary experiment in \secref{subsec:basic-eval}.
}%
{%
  まず, Type AとType Bのネットワークにおける学習型フィードバック制御を実行し, ネットワーク構造間における精度の比較を行う.
  次に, Type BにおけるLossの定義による動作の違い, 姿勢や重量物体を把持したとき等の状況に応じた学習型フィードバックの動作の違いを比較する.
  最後に, 学習型制御の定量的評価と, 提案手法と視覚を用いた捺印に関して実験を行う.

  本章においては主に, \secref{subsec:basic-eval}と同様に, ランダムな関節角度$\bm{\theta}_{rand}$から指令関節角度$\bm{\theta}^{ref}$を実現する実験を行う.
  初めに$\bm{l}^{ref} = \bm{h}(\bm{\theta}^{ref}, \bm{f}_{const})$によりある程度関節角度を実現し, それを施行$N_{trial}=0$回目として\secref{subsec:learning-feedback}で説明したフィードバックを行う度に$||\bm{\theta}^{ref}-\bm{\theta}^{cur}||_{2}$を計測する.
  $\bm{\theta}^{cur}$から$\bm{\theta}^{ref}$まで$N^{max}_{trial}$回($0 \leq N_{trial} < N^{max}_{trial}$)のフィードバックを行うことを, 初期関節角度$\bm{\theta}_{rand}$を変えながら$N^{max}_{rand}$回($0 \leq N_{rand} < N^{max}_{rand}$)繰り返し, $N_{trial}$回目における$N^{max}_{rand}$回分の$||\bm{\theta}^{ref}-\bm{\theta}^{cur}||_{2}$の平均と分散の遷移を確認する.
  このグラフを, $\textrm{Graph}_{ave}(\bm{\theta}^{ref}, N^{max}_{trial}, N^{max}_{rand})$と表す.
  また, 平均と分散を取らず$N^{max}_{rand}$回分の$||\bm{\theta}^{ref}-\bm{\theta}^{cur}||_{2}$の遷移を表したグラフを$\textrm{Graph}_{raw}(\bm{\theta}^{ref}, N^{max}_{trial}, N^{max}_{rand})$と表す.
}%

\begin{figure}[t]
  \centering
  \includegraphics[width=0.9\columnwidth]{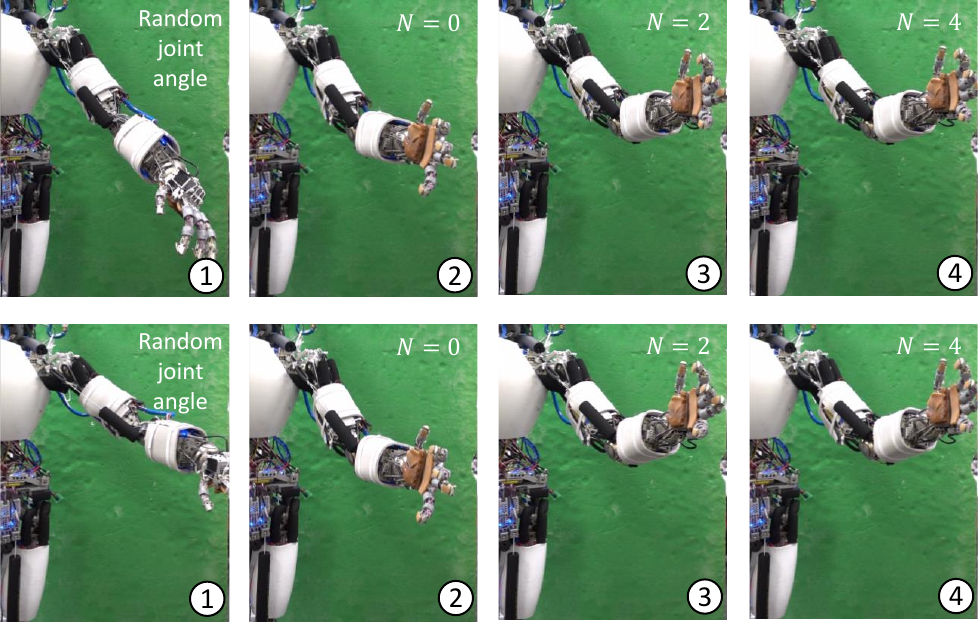}
  \vspace{-1.0ex}
  \caption{The experiment of online learning feedback control.}
  \label{figure:experimental-appearance}
  \vspace{-3.0ex}
\end{figure}

\begin{figure}[t]
  \centering
  \includegraphics[width=0.95\columnwidth]{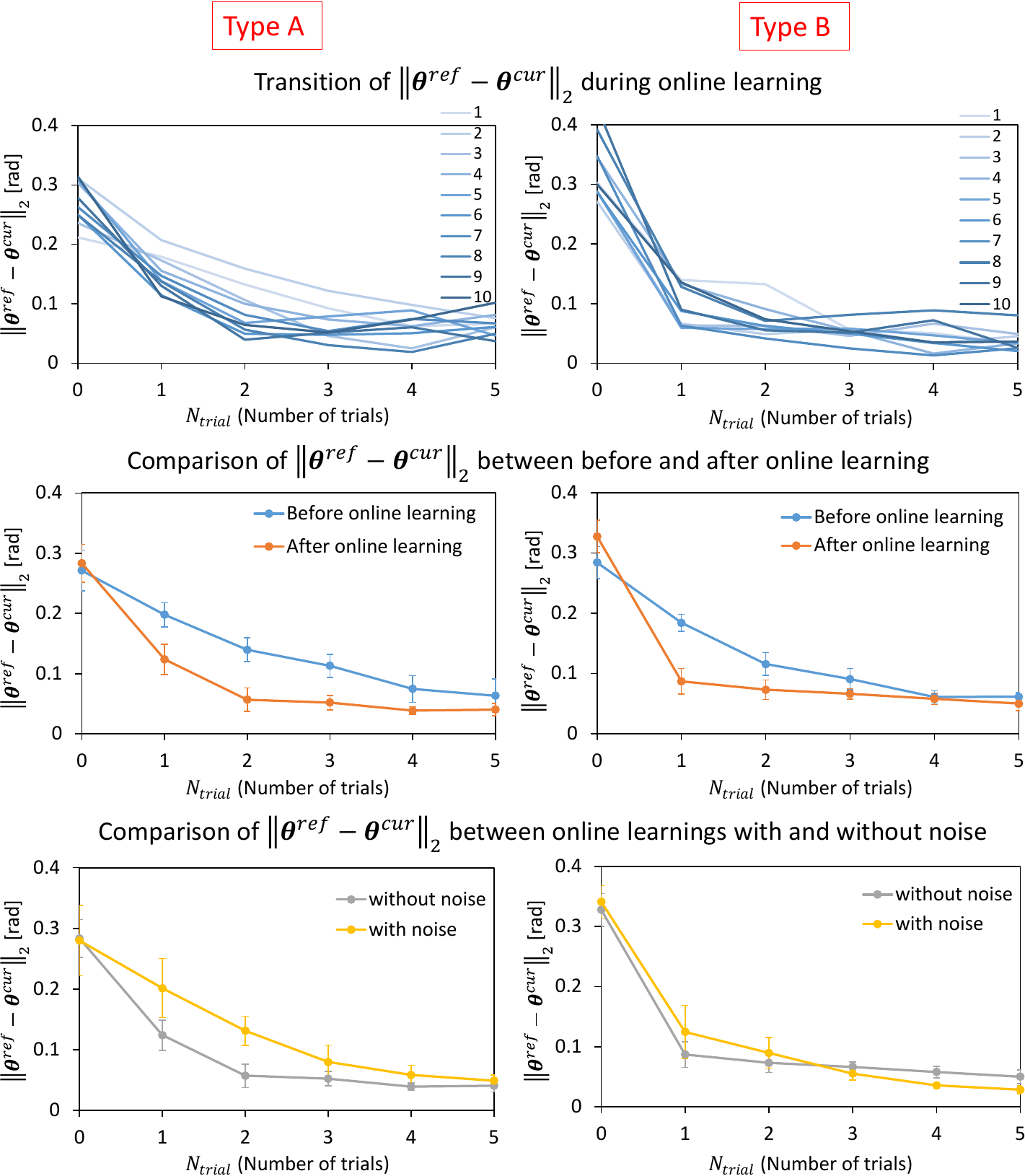}
  \vspace{-1.0ex}
  \caption{Comparison of online learning feedback controls between using the networks of Type A and Type B.}
  \label{figure:normal-vs-compare}
  \vspace{-3.0ex}
\end{figure}


\subsection{Learning Feedback Control with Different Network Structures} \label{subsec:different-structure}
\switchlanguage%
{%
  By setting $\bm{\theta}^{ref}$ as a certain joint angle $\bm{\theta}_{0} = (\theta_{S-r}, \theta_{S-p}, \theta_{S-y}, \theta_{E-p}, \theta_{E-y}) = (30, -30, 30, -60, 30)$ [deg], we compare OLFC using the networks of Type A and B in the same experiment.

  We show the experiment in \figref{figure:experimental-appearance} and the experimental results in \figref{figure:normal-vs-compare}.
  In the upper graphs of \figref{figure:normal-vs-compare}, we show $\textrm{Graph}_{raw}(\bm{\theta}^{ref}=\bm{\theta}_{0}, N^{max}_{trial}=5, N^{max}_{rand}=10)$ during online learning stated in \secref{subsec:online-learning}.
  While $||\bm{\theta}^{ref}-\bm{\theta}^{cur}||_{2}$ regarding Type A gradually decreases as $N_{rand}$ increases, the online learning regarding Type B converges after a few $N_{rand}$.

  The middle graphs of \figref{figure:normal-vs-compare} show $\textrm{Graph}_{ave}(\bm{\theta}^{ref}=\bm{\theta}_{0}, N^{max}_{trial}=5, N^{max}_{rand}=5)$ before and after the online learning.
  In subsequent experiments, the online learning procedure is not working when conducting comparison experiments.
  We can see that the convergence of joint angles is improved very much regarding both Type A and B.
  $||\bm{\theta}^{ref}-\bm{\theta}^{cur}||_{2}$ regarding Type B converges more quickly than Type A.

  Finally, we discuss the difference between Type A and B, and conduct a comparative experiment.
  The transition of joint angles in the network of Type B holds true at all times but it does not necessarily hold regarding Type A.
  In the case of the redundant flexible musculoskeletal structure, we can consider various $\Delta\bm{l}$ to realize $\bm{\theta}^{ref}$ from $\bm{\theta}^{cur}$, and the result of the online learning should depend on the training data.
  So we conduct the same experiments as stated above with added noise to the data for online learning, as shown below,
  \begin{align}
    \Delta\bm{l} \gets \Delta\bm{l} + \textrm{Random}(a, b)
  \end{align}
  where $\textrm{Random}(a, b)$ is the random value in $[a, b]$ ($a=0, b=5$ [mm] in this study).
  We show $\textrm{Graph}_{ave}(\bm{\theta}^{ref}=\bm{\theta}_{0}, N^{max}_{trial}=5, N^{max}_{rand}=5)$ with or without the additional noise during the online learning in the lower graphs of \figref{figure:normal-vs-compare}.
  While the result of the online learning does not vary with or without the additional noise regarding Type B, the convergence largely deteriorates regarding Type A and its variance is comparatively large.

  We use the network of Type B in remaining experiments.
}%
{%
  $\bm{\theta}^{ref}$を$\bm{\theta}_{0} = (\theta_{S-r}, \theta_{S-p}, \theta_{S-y}, \theta_{E-p}, \theta_{E-y}) = (30, -30, 30, -60, 30)$ [deg]として, Type AとType Bにおいて全く同じ実験を行い比較する.

  実験の様子を\figref{figure:experimental-appearance}に, 実験結果を\figref{figure:normal-vs-compare}に示す.
  \figref{figure:normal-vs-compare}の上図に, オンライン学習の過程における, $\textrm{Graph}_{raw}(\bm{\theta}^{ref}=\bm{\theta}_{0}, N^{max}_{trial}=5, N^{max}_{rand}=10)$を表す.
  Type Aにおいては$N_{rand}$ごとに徐々に$||\bm{\theta}^{ref}-\bm{\theta}^{cur}||_{2}$の遷移グラフが下がっていくのに対して, Type Bでは$N_{rand}=1, 2$回程度で学習が収束していることがわかる.

  また, \figref{figure:normal-vs-compare}の中図は, 学習前と学習後における$\textrm{Graph}_{ave}(\bm{\theta}^{ref}=\bm{\theta}_{0}, N^{max}_{trial}=5, N^{max}_{rand}=5)$を表している.
  Type A, Type Bどちらにおいても, 学習前に比べて大きく収束が改善していることがわかる.
  違いとして, Type Aの方がType Bに比べてなだらかに$||\bm{\theta}^{ref}-\bm{\theta}^{cur}||_{2}$が遷移していることがわかる.

  最後に, これらのType A, Type Bの違いについて考察し比較実験を行う.
  これらの違いは, Type Bは確実に成り立つ遷移なのに対して, Type Aは成り立つとは限らないところである.
  冗長で柔軟な筋骨格構造を持つ場合は, ある$\bm{\theta}^{cur}$を$\bm{\theta}^{ref}$まで動作させる$\Delta\bm{l}$は複数考えられ, その学習結果は実際の学習データに依存してしまうはずである.
  そこで, 以下のようにフィードバック動作時の値にノイズを加え, 同様に実験を行った.
  \begin{align}
    \Delta\bm{l} = \Delta\bm{l} + \textrm{Random}(a, b)
  \end{align}
  ここで, $\textrm{Random}(a, b)$は, $[a, b]$の間のランダムな値を表す(本研究では$a=0, b=5$ [mm]とした).
  オンライン学習を実行する際にノイズを加えた時と加えないときの$\textrm{Graph}_{ave}(\bm{\theta}^{ref}=\bm{\theta}_{0}, N^{max}_{trial}=5, N^{max}_{rand}=5)$を\figref{figure:normal-vs-compare}の下図に示す.
  Type Bのときはノイズを加えないときと加えたときで大きく学習結果が異ならないのに対して, Type Aでは著しく学習結果が悪くなっており, 分散も比較的大きいことがわかる.

  これらから, Type Aに比べType Bのネットワークはよりオンライン学習に適し, $L$の設計等より発展が見込めるため, 本研究の残りの実験ではType Bを用いて実験を行う.
}%

\begin{figure}[t]
  \centering
  \includegraphics[width=0.95\columnwidth]{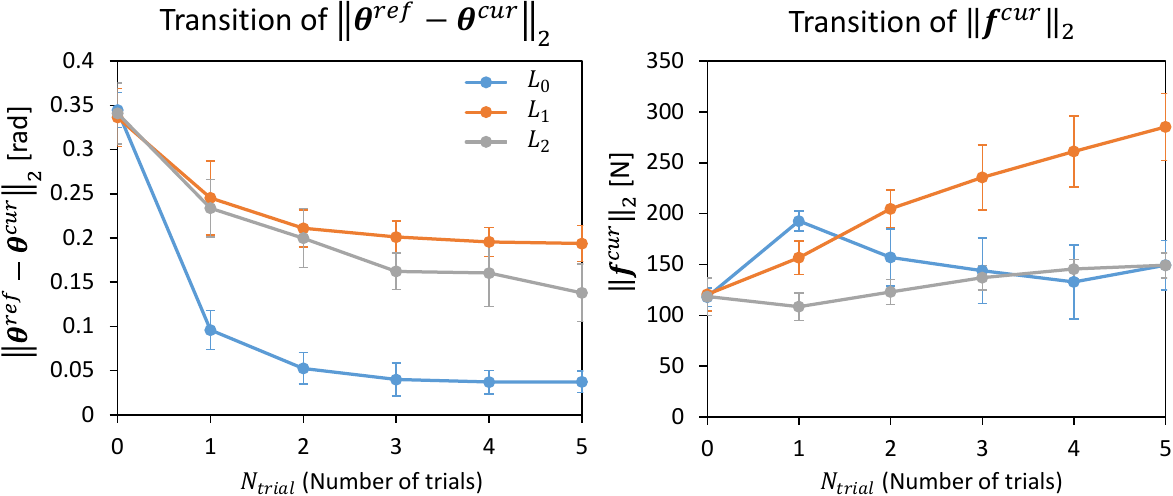}
  \vspace{-1.0ex}
  \caption{Comparison of online learning feedback controls among various loss definitions.}
  \label{figure:various-loss}
  \vspace{-1.0ex}
\end{figure}

\subsection{Learning Feedback Control with Various Loss Definition} \label{subsec:various-loss}
\switchlanguage%
{%
  As stated in \secref{subsec:learning-feedback}, regarding Type B, we can change the behavior of OLFC by changing the definition of $L$.
  We show the difference of $\textrm{Graph}_{ave}(\bm{\theta}^{ref}=\bm{\theta}_{0}, N^{max}_{trial}=5, N^{max}_{rand}=5)$ among the loss definitions of $L_0$ (\equref{eq:loss0}), $L_1$ (\equref{eq:loss1}), and $L_2$ (\equref{eq:loss2}) in the left graph of \figref{figure:various-loss}.
  In the right graph of \figref{figure:various-loss}, we show $||\bm{f}^{cur}||_{2}$ instead of $||\bm{\theta}^{ref}-\bm{\theta}^{cur}||_{2}$ regarding $\textrm{Graph}_{ave}$ ($\bm{f}^{cur}$ is the current muscle tension).
  We can see that the results using $L_1$ and $L_2$ deteriorate compared to $L_0$, due to the additional limitation of $\Delta\bm{l}$.
  Also, the result of $L_2$ is better than that of $L_1$.
  Regarding $||\bm{f}^{cur}||_{2}$, while the muscle tension gradually increases when using $L_1$, the muscle tension does not vary much when using $L_2$.
  When using $L_0$, the muscle tension increases in the beginning, but gradually decreases afterwards.
  However, the variance of the muscle tension is large, and so high muscle tension sometimes emerges.
}%
{%
  \secref{subsec:learning-feedback}で述べたように, Type Bにおいては$L$の設計によって挙動を変えることができる.
  $L$としてLoss0 (\equref{eq:loss0}), Loss1 (\equref{eq:loss1}), Loss2 (\equref{eq:loss2})を用いた時の$\textrm{Graph}_{ave}(\bm{\theta}^{ref}=\bm{\theta}_{0}, N^{max}_{trial}=5, N^{max}_{rand}=5)$の違いを\figref{figure:various-loss}の左図に示す.
  \figref{figure:various-loss}の右図には, $\textrm{Graph}_{ave}$と同様だが, $||\bm{\theta}^{ref}-\bm{\theta}^{cur}||_{2}$ではなく, $||\bm{f}^{cur}||_{2}$を表示している($\bm{f}^{cur}$は現在の筋張力を表す).
  Loss1, Loss2における結果は, $\Delta\bm{l}$に関する制約を加えているため, Loss0に比べると大きく精度が落ちることがわかる.
  また, Loss2はLoss1に比べると精度が良い.
  $||\bm{f}^{cur}||_{2}$に関しては, Loss1では徐々に筋張力が高まるのに対して, Loss1ではほとんど筋張力に変化が無いことがわかる.
  また, Loss0では始めは筋張力が高まり, その後下がっていくが, 分散が大きく, 大きな筋張力が発生してしまうこともある.
}%

\begin{figure}[t]
  \centering
  \includegraphics[width=0.95\columnwidth]{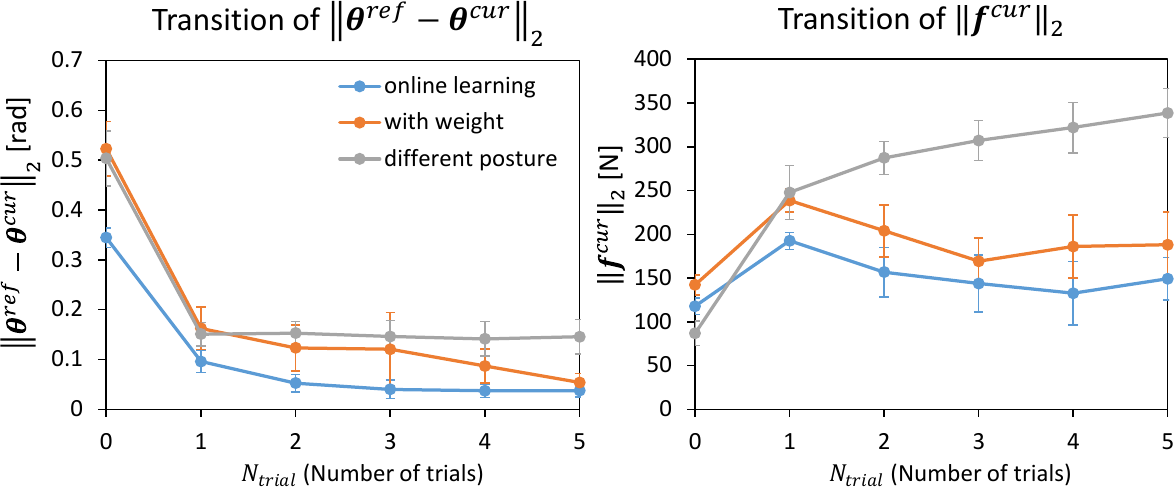}
  \vspace{-1.0ex}
  \caption{Comparison of online learning feedback controls among various situations.}
  \label{figure:various-situation}
  \vspace{-3.0ex}
\end{figure}

\subsection{Learning Feedback Control in Various Situations} \label{subsec:various-situation}
\switchlanguage%
{%
  We verify whether the online learning result is valid in situations different from those when conducting the online learning.
  As in \secref{subsec:different-structure}, we conducted the online learning by setting $N^{max}_{trial}=5, N^{max}_{rand}=10$.
  We verified the difference of $\textrm{Graph}_{ave}(\bm{\theta}^{ref}=\bm{\theta}_{0}, N^{max}_{trial}=5, N^{max}_{rand}=5)$ among the state with online learning, the state grasping a heavy object (3.6 kg), and the state with a forward-bent posture (45 deg).
  During these experiments, the online learning was not executed.
  As with \secref{subsec:various-loss}, we also checked $||\bm{f}^{cur}||_{2}$.
  The result is shown in \figref{figure:various-situation}.
  By grasping the heavy object or changing the posture, $||\bm{\theta}^{ref}-\bm{\theta}^{cur}||_{2}$ deteriorates at $N_{trial}=0$, because the space of muscle tension which is not trained by \cite{kawaharazuka2018bodyimage} is used.
  However, even with changes in situation, the convergence is faster than before the online learning.
  In the left graph of \figref{figure:various-situation}, we can see that the muscle tension increases as a whole when grasping the heavy object.
  Also, when changing the posture, the joint torque is insufficient, high muscle tension emerges, and the convergence stops at a certain value.
}%
{%
  オンライン学習を行った状況とは異なった状況で学習結果が有用かどうかを検証する.
  \secref{subsec:different-structure}と同様に$N^{max}_{trial}=5, N^{max}_{rand}=10$でオンライン学習後, オンライン学習を止める.
  学習を行った際と同じ状態, 重量物体(3.6 kg)を把持した状態, 姿勢を変えた状態 (体幹を45度前傾), においてどのように$\textrm{Graph}_{ave}(\bm{\theta}^{ref}=\bm{\theta}_{0}, N^{max}_{trial}=5, N^{max}_{rand}=5)$が変化するかを検証する.
  \secref{subsec:various-loss}と同様に, $||\bm{f}^{cur}||_{2}$についても検証する.
  結果を\figref{figure:various-situation}に示す.
  重量物体の把持や姿勢の変更により, \cite{kawaharazuka2018bodyimage}では学習されていない筋張力の空間を使っているため$N_{trial}=0$における精度は大きく下がっている.
  しかし, 重量物体を持った際や姿勢を変えた際でも, 学習前のなだらかな収束に比べて収束の速度はかなり速い.
  \figref{figure:various-situation}の右図からは, 重量物体を持った際に筋張力が全体的に高まっていること, 姿勢を変えた実験の際には筋張力が足りず大きな力が出てしまい, 左図の収束も一定で止まっていることがわかる.
}%

\begin{figure}[t]
  \centering
  \includegraphics[width=0.95\columnwidth]{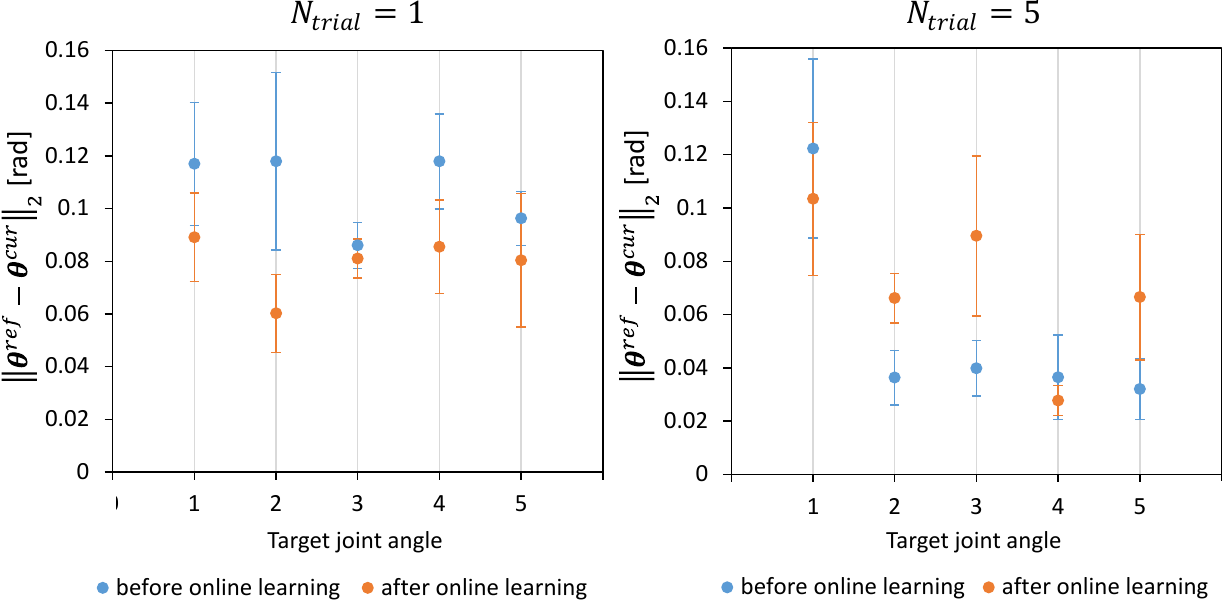}
  \vspace{-1.0ex}
  \caption{Quantitative evaluation of online learning feedback control.}
  \label{figure:quantitative-eval}
  \vspace{-3.0ex}
\end{figure}

\subsection{Quantitative Evaluation of Learning Feedback Control} \label{subsec:quantitative-evaluation}
\switchlanguage%
{%
  So far, we have verified OLFC regarding only one certain joint angle by setting $\bm{\theta}^{ref}=\bm{\theta}_{0}$.
  In this experiment, we will evaluate OLFC quantitatively for movements that realize the randomly chosen various joint angles.
  First, we randomly generated 5 target joint angles $\bm{\theta}^{ref}$ for evaluation.
  We conducted the feedback control experiment to realize each $\bm{\theta}^{ref}$ by setting $N^{max}_{trial}=5$ and $N^{max}_{rand}=5$.
  Second, we conducted the feedback control experiment to realize the random joint angle $\bm{\theta}'_{rand}$ from the random joint angle $\bm{\theta}_{rand}$ by setting $N^{max}_{trial}=5$ and $N^{max}_{rand}=20$, while running the online learning.
  Finally, we conducted the same experiment as before online learning to realize each $\bm{\theta}^{ref}$ by setting $N^{max}_{trial}=5$ and $N^{max}_{rand}=5$.

  We show the average and variance of $||\bm{\theta}^{ref}-\bm{\theta}^{cur}||_{2}$ of $N^{max}_{rand}$ times at $N_{trial}=1$ and $N_{trial}=5$ regarding each $\bm{\theta}^{ref}$ before and after the online learning in \figref{figure:quantitative-eval}.
  Regarding $N_{trial}=1$, the precision after the online learning is better than before the online learning at any $\bm{\theta}^{ref}$, even if $\bm{\theta}^{ref}$ is not included in $\bm{\theta}'_{rand}$.
  On the other hand, regarding $N_{trial}=5$, the precisions before and after the online learning do not largely vary.
  There are many cases where the variance before the online learning is better than that after the online learning.
}%
{%
  これまでは$\bm{\theta}^{ref}=\bm{\theta}_{0}$としてある一つの関節角度に対して検証してきた.
  本実験では, ランダムな関節角度$\bm{\theta}_{rand}$に関して動作をさせることで提案手法を定量的に評価する.
  まず, 評価する関節角度$\bm{\theta}^{ref}$をランダムに5つ生成する.
  これらに対してこれまでと同様に$N^{max}_{trial}=5, N^{max}_{rand}=5$でフィードバック実験を行う.
  次に, ランダムな関節角度$\bm{\theta}_{rand}$からランダムな関節角度$\bm{\theta}'_{rand}$に$N^{max}_{trial}=5$でフィードバックする実験を$N^{max}_{rand}=20$回行い, その際にオンラインで学習をしていく.
  最後に, 学習前と同様の$\bm{\theta}^{ref}$に対して同様に$N^{max}_{trial}=5, N^{max}_{rand}=5$でフィードバック実験を行う.

  $N_{trial}=1, N_{trial}=5$の際における$||\bm{\theta}^{ref}-\bm{\theta}^{cur}||_{2}$の平均と分散を, それぞれの$\bm{\theta}^{ref}$について\figref{figure:quantitative-eval}に示す.
  $N_{trial}=1$においては, 学習後は学習前に比べてどの$\bm{\theta}^{ref}$についても精度が改善していることがわかる.
  つまり, $\bm{\theta}'_{rand}$に$\bm{\theta}^{ref}$が含まれていなくても, オンライン学習は汎化することがわかった.
  一方, $N_{trial}=5$においては, 学習前と学習後の精度はあまり変わらず, 分散が小さいという観点からは学習をしない場合のほうが精度が良い場合が多い.
}%

\subsection{Japanese Stamp Pushing Experiment} \label{subsec:stamp-pushing}
\begin{figure}[t]
  \centering
  \includegraphics[width=0.85\columnwidth]{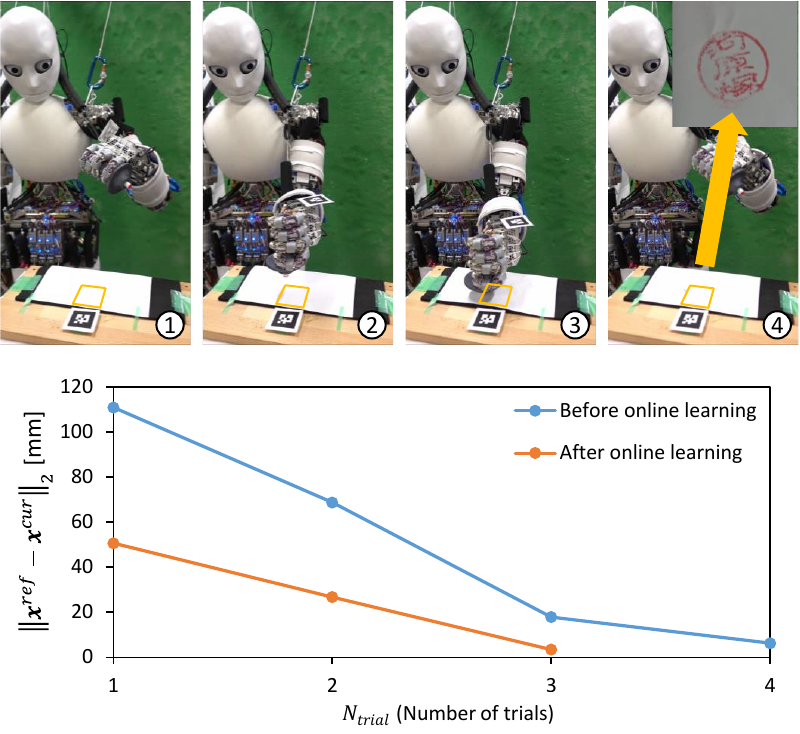}
  \vspace{-1.0ex}
  \caption{Experiment of pushing a Japanese stamp by visual feedback.}
  \label{figure:hanko}
  \vspace{-3.0ex}
\end{figure}

\switchlanguage%
{%
  We will conduct a practical task of pushing a Japanese stamp, integrating the proposed method and visual information.
  We put an AR marker on the desk, and the robot pushes a Japanese stamp at the position $\bm{x}_{place}$ which is 50 mm from the marker toward the robot.
  Also, as stated in \secref{subsec:basic-structure}, we estimated the actual joint angle from the vision sensor using the method of \cite{kawaharazuka2018online}.
  This is a method to estimate the actual joint angle $\bm{\theta}^{est}$ by attaching an AR marker to the hand, setting the estimated joint angle from muscle length as the initial joint angle, and solving inverse kinematics to the marker.
  Since the link length and joint position of the geometric model are not necessarily correct, we can conduct the position alignment more precisely by using $\bm{\theta}^{est}$.
  The current position of the Japanese stamp is expressed as $\bm{x}^{cur}$ calculated from the obtained $\bm{\theta}^{est}$.
  Similarly, by setting $\bm{\theta}^{est}$ as the initial joint angle and solving inverse kinematics to $\bm{x}^{ref}$ (we will explain $\bm{x}^{ref}$ below), we obtain the target joint angle $\bm{\theta}^{ref}$.
  In this experiment, we do not conduct the movement at $N_{trial}=0$ as in previous experiments, but conduct only the feedback control to realize $\bm{\theta}^{ref}$ from $\bm{\theta}^{cur}=\bm{\theta}^{est}$, and evaluate $||\bm{x}^{ref}-\bm{x}^{cur}||_{2}$ at each step.
  In the beginning, the robot with a Japanese stamp moves its arm to 50 mm above $\bm{x}_{place}$ by OLFC, and after that, sets the stamp down to $\bm{x}_{place}$.
  Finally, by changing the rotation of the arm at $\bm{x}_{place}$ several times, the robot firmly pushes the Japanese stamp.

  We show the experiment and transition of $||\bm{x}^{ref}-\bm{x}^{cur}||_{2}$ in \figref{figure:hanko}.
  In this experiment, $\bm{x}^{ref}$ is the position 50 mm above $\bm{x}_{place}$.
  The case after online learning shows a better convergence than before online learning.
  Although the stamp was firmly pushed, the stamp contacted the paper before convergence when moving to $\bm{x}_{place}$ from $\bm{x}^{ref}$, and it was difficult to push the stamp at the exact place.
}%
{%
  最後に, 視覚情報と本研究を合わせることで, 捺印する実験を行う.
  机の上にARマーカを置き, 50 mm手前$\bm{x}_{place}$に印鑑を押す.
  また, \secref{subsec:basic-structure}で述べたように, \cite{kawaharazuka2018online}の手法を用いて視覚から実機の関節角度情報を得る.
  これは, 手先にARマーカを貼り付け, 筋長変化による関節角度推定値を初期値として, そのマーカに対して逆運動学を解くことで, 実機関節角度$\bm{\theta}^{est}$を推定する方法である.
  リンク長や関節位置等の幾何モデルは正しいとは限らないため, この$\bm{\theta}^{est}$を用いることで, より正確に位置合わせをすることができる.
  この$\bm{\theta}^{est}$から得られる印鑑の現在位置を$\bm{x}^{cur}$とする.
  同様に, $\bm{\theta}^{est}$を初期値として, 後に述べる$\bm{x}^{ref}$に対して逆運動学を解くことで, 指令関節角度$\bm{\theta}^{ref}$を求める.
  本実験では, これまでの実験における$N_{trial}=0$の動作は行わず, \secref{subsec:learning-feedback}で述べた$\bm{\theta}^{cur}=\bm{\theta}^{est}$から$\bm{\theta}^{ref}$に対するフィードバックを行い, その度に$||\bm{x}^{ref}-\bm{x}^{cur}||_{2}$を評価する.
  実験としては, ロボットが印鑑を持った状態から始め, $\bm{x}_{place}$の50 mm上までフィードバックを用いて動作し, その後$\bm{x}_{place}$に対してフィードバックを用いて同様に動作をする.
  最後に, $\bm{x}_{place}$に対する角度変えて何度か動作させることで, しっかりと捺印する.

  実験の様子と$||\bm{x}^{ref}-\bm{x}^{cur}||_{2}$の遷移を\figref{figure:hanko}に示す.
  ここでは, $\bm{x}^{ref}$は最初の動作である印鑑を押す位置$\bm{x}_{place}$の50 mm上である.
  オンライン学習前と学習後では, これまでと同様学習後の方が良い収束を示している.
  また, 印鑑はしっかりと押すことができたものの, $\bm{x}^{ref}$から$\bm{x}_{place}$まで動かす際, 収束する前に印鑑が紙と接触してそれ以上動かなくなり, 完全に正確な位置に対して捺印をすることはできていない.
}%

\section{Discussion} \label{sec:discussion}
\subsection{Experimental Results}
\switchlanguage%
{%
  In \secref{subsec:different-structure}, we found that the convergence of online learning is faster for Type B, which represents the transition of the joint angle, than for Type A, which calculates the target change in muscle length from two joint angles.
  After online learning, the error of the feedback control drops gently for Type A, while it drops sharply for Type B.
  Also, the performance of Type A deteriorates markedly when the noise is added to the feedback control, while the performance of Type B does not change at all.
  This means that the convergence of the model by online learning and the convergence of the joint angle error by the feedback control using the learned model are better for Type B than for Type A.
  It is important to note that Type A does not necessarily hold true while Type B holds true at all times, which may contribute significantly to the performance of the network.

  In \secref{subsec:various-loss}, we found that the loss of minimizing $\Delta\bm{l}$ prevents the antagonist muscles from loosening and increases the internal force in $L_1$.
  In contrast, $L_2$ avoids the increase in muscle tension compared to $L_0$, although its accuracy is lower.
  Therefore, $L_0$ is useful when accuracy is emphasized and $L_2$ is also useful when the increase in muscle tension should be avoided.

  In \secref{subsec:various-situation}, we found that the convergence of the joint angle errors after learning is faster than that before learning and that this method is useful, even in different situations.

  In \secref{subsec:quantitative-evaluation}, we found that the accuracy after online learning is better for $N_{trial}=1$ than before online learning, while there is no significant difference in the accuracy before and after the learning for $N_{trial}=5$.
  We can consider that this is because learning globally is easy, but the online learning is difficult when $||\bm{\theta}^{cur}-\bm{\theta}^{ref}||_{2}$ is small.
  Therefore, it is important to use this method for a very small number of feedback trials.

  In \secref{subsec:stamp-pushing}, we found that the position error after learning is better converged than before learning.
  Visual feedback can be constructed by combining visual information and the proposed feedback control, and that this study can be applied to musculoskeletal humanoids without joint angle sensors, using $\bm{\theta}^{est}$.
}%
{%
  実験に関する考察を行う.

  \secref{subsec:different-structure}ではType AとType Bのネットワークについてフィードバック制御のパフォーマンスを調べた.
  実験結果から, Type Bの方がType Aに比べてonlne learningの収束のスピード速いことがわかった.
  学習後のネットワークを用いて比較すると, フィードバック制御においてType Aではなだらかに誤差が下がっていくのに対して, Type Bでは急激に誤差が下がることがわかった.
  また, フィードバック時にノイズを加えるとType Aは顕著に性能が悪くなるのに対して, Type Bでは性能はほとんど変化しなかった.
  これらから, Type Aに対してType Bの方がonline learningによるモデルの収束性, 学習されたモデルを使ったフィードバックによる関節角度誤差の収束性が良いと考えられる.
  Type Bは確実に成り立つ遷移なのに対して, Type Aは成り立つとは限らない, という性質は重要であり, これがネットワークの性能に大きな寄与を与えている可能性が大きい.

  \secref{subsec:various-loss}では, Type Bにおける特徴であるLossの変化によるフィードバック制御の挙動の変化について調べた.
  実験結果から,  Loss1は$\Delta\bm{l}$を最小化することで拮抗筋が緩まなくなり, 内力が高まってしまうことがわかった.
  これに対して, Loss2は精度は落ちるもののLoss0に比べて筋張力の高まりを回避することができる.
  ゆえに, Type Bにおいて, 精度を重視する場合にはLoss0が有用であり, 筋張力の高まりを回避したい場合はLoss2も有用であることがわかった.

  \secref{subsec:various-situation}では, 重要物を持った時, 姿勢を変えた時, におけるフィードバック制御の挙動を調べた.
  実験結果から, それら状況を変化させても学習前に比べて関節角度誤差の収束は速く, 学習時と異なる状況おいても本手法が有用であることがわかった.

  \secref{subsec:quantitative-evaluation}ではランダムな関節角度に対する本手法の効果を定量的に評価した.
  実験結果から, $N_{trial}=1$のときは学習前に比べて精度が良くなるものの, $N_{trial}=5$のときは学習前と後において大きく精度が変わらないことがわかった.
  これは, 大域的な学習は簡単なものの, $||\bm{\theta}^{cur}-\bm{\theta}^{ref}||_{2}$が小さい場合は学習が難しいことが理由であると考えられる.
  よって, 本手法は非常に少ない回数のフィードバック試行で用いることが重要であることがわかる.

  \secref{subsec:stamp-pushing}では資格情報と本フィードバック制御を用いて捺印を行った.
  実験結果から, 学習後は学習前に比べてより良い収束結果を示している.
  よって, 視覚情報と本研究を合わせることでvisual feedbackが構築可能であり, $\bm{\theta}^{est}$を用いることから, 本研究は関節角度センサのない筋骨格ヒューマノイドに対しても適用可能なことがわかった.

}%

\subsection{Limitations and Future Works}
\switchlanguage%
{%
  Our method has the following limitations: (i) it assumes quasi-static behavior, and (ii) it does not completely reduce the joint angle error to zero.

  As for (i), if the robot moves at high speed, the assumption of the network does not hold and the proposed method may not be applied well.
  On the other hand, although dynamic effects can be considered by the recurrent network structure at high frequency, there are some problems such as a large amount of data and difficulty in learning a large space of time series data.
  Therefore, it is necessary to develop more efficient sampling and learning methods.

  As for (ii), this feedback control does not realize the target joint angle as accurately as the feedback control in the axis-driven type.
  This characteristic is important, and we realize that a new paradigm of control for musculoskeletal humanoids is required, as well as the accurate realization of the joint angle as in this study.
  We would like to establish a control method to accomplish the task by making full use of various environments and tools, no matter how much the joint angle error is.
}%
{%
  本研究の手法は(i)準静的動作を仮定している(ii)誤差を完全にゼロにするものではない, という制限がある.

  (i)については, ロボットが速いスピードで動く場合には, ネットワークの前提が崩れ, 上手く適用できない可能性がある.
  これに対して, 動的な影響を考慮する手法は速い周期でRNNなどのモデルを使えば可能であるが, データが膨大になる, 広い空間の学習が難しいなどの問題を孕む.
  そのため, より効率の良いサンプリングと学習手法を構築していく必要がある.

  (ii)については, 本研究の実験結果から分かったように, 軸駆動型ほどフィードバック制御によって精度良く指令関節角度が実現できるわけではないということである.
  この性質は重要であり, 本研究のようなある程度正確な関節角度実現は必要であるとともに, 全く新しいパラダイムの制御が必要であるとも実感される.
  誤差を許容してでも様々な環境や道具を駆使してタスクを達成する制御手法が今後強く求められると考える.
}%

\section{CONCLUSION} \label{sec:conclusion}
\switchlanguage%
{%
  In this study, we proposed an online learning feedback control system considering hysteresis of musculoskeletal structures, and compared its network structures and loss definitions.
  By constructing a network that represents the transition of the current joint angle by the change in muscle length, we can calculate the optimized change in muscle length to realize the target joint angle through backpropagation and gradient descent.
  The convergence of the feedback control using this network is better than that of using the network with the current and target joint angles as the input and the change in muscle length as the output.
  We can inhibit high internal muscle tension by changing the definition of the loss function.
  Also, this method can work in situations different from those of the online learning, such as with external force or with different posture, and is shown to have a generalization ability from the quantitative experiment.
  This method can be applied to the musculoskeletal humanoid without joint angle sensors, and we can conduct visual feedback by using vision sensor.
}%
{%
  本研究では, 筋構造構造におけるヒステリシスを考慮したオンライン学習型フィードバック制御について考案し, ネットワーク構造や計算方法等に関して比較・考察をした.
  ある筋長変化による現在関節角度の遷移を記述するニューラルネットワークを構成し, 誤差逆伝播から指令関節角度を実現する最適な筋長変化を計算することができる.
  この手法は, 関節角度と目標関節角度から筋長変化を求める手法よりもオンライン学習結果やフィードバックの収束結果が良く, 誤差逆伝播の際の損失関数の設計により内力の高まりを抑えることも可能である.
  また, 本手法は外力が働いたり姿勢が変わったり等, 学習時と違う状況でも一定の効果を発揮し, 定量的評価により汎化性能があることもわかった.
  本手法は関節角度を測ることができない筋骨格ヒューマノイドに対しても適用可能であり, 視覚と組み合わせることでvisual feedbackを行うこともできる.

  今後はより高度なマニピュレーション, 動的な動作等に取り組みたい.
}%

{
  \bibliographystyle{IEEEtran}
  \bibliography{main}
}

\end{document}